\documentclass[conference]{IEEEtran}
\IEEEoverridecommandlockouts
\usepackage{cite}
\usepackage{amsmath,amssymb,amsfonts}
\usepackage{graphicx}
\usepackage{textcomp}
\usepackage{xcolor}
\def\BibTeX{{\rm B\kern-.05em{\sc i\kern-.025em b}\kern-.08em
    T\kern-.1667em\lower.7ex\hbox{E}\kern-.125emX}}

\usepackage{algorithm}
\usepackage{algorithmicx}
\usepackage[noend]{algpseudocode}
\usepackage[caption=false,font=normalsize,labelfont=sf,textfont=sf]{subfig}

\DeclareMathOperator*{\argmin}{argmin}
\begin{document}

\title{Incremental Learning with Concept Drift Detection and Prototype-based Embeddings for Graph Stream Classification
\thanks{This paper was supported by the European Research Council (ERC) under grant agreement No 951424 (Water-Futures), the European Union’s Horizon 2020
research and innovation programme under grant agreement No 739551 (KIOS CoE), the Republic of Cyprus through the Deputy Ministry of Research, Innovation and Digital Policy, and the Cyprus Academy of Sciences, Letters and Arts.}
}



\author{
	\IEEEauthorblockN{
		Kleanthis Malialis\textsuperscript{a},
		Jin Li\textsuperscript{a, b},
		Christos G. Panayiotou\textsuperscript{a, b} and
		Marios M. Polycarpou\textsuperscript{a, b}
	}
	\IEEEauthorblockA{
		\textsuperscript{a} \textit{KIOS Research and Innovation Center of Excellence}\\
		\textsuperscript{b} \textit{Department of Electrical and Computer Engineering}\\
		\textit{University of Cyprus},
		Nicosia, Cyprus\\
		Email: \{malialis.kleanthis, li.jin, christosp, mpolycar\}@ucy.ac.cy\\
		ORCID: \{0000-0003-3432-7434, 0000-0002-3534-524X, 0000-0002-6476-9025, 0000-0001-6495-9171\}
	}
}

\maketitle

\begin{abstract}
Data stream mining aims at extracting meaningful knowledge from continually evolving data streams, addressing the challenges posed by nonstationary environments, particularly, concept drift which refers to a change in the underlying data distribution over time. Graph structures offer a powerful modelling tool to represent complex systems, such as, critical infrastructure systems and social networks. Learning from graph streams becomes a necessity to understand the dynamics of graph structures and to facilitate informed decision-making. This work introduces a novel method for graph stream classification which operates under the general setting where a data generating process produces graphs with varying nodes and edges over time. The method uses incremental learning for continual model adaptation, selecting representative graphs (prototypes) for each class, and creating graph embeddings. Additionally, it incorporates a loss-based concept drift detection mechanism to recalculate graph prototypes when drift is detected.
\end{abstract}


\section{Introduction}\label{sec:intro}
The explosive growth of data that we currently encouter has rendered traditional data analysis methods insufficient in addressing the velocity, volume, and variety of data generated. Data stream mining is dedicated to extracting insights and meaningful knowledge from high-speed, continuously evolving data streams. These time-varying and non-stationary environments pose significant challenges as the underlying data distribution evolves over time, which is referred to as concept drift \cite{ditzler2015learning, gama2014survey, lu2018learning}.

Concept drift can manifest itself in various forms; for example, in critical infrastructure systems \cite{kyriakides2014intelligent} (such as, water networks, power and energy systems, transportation networks, and telecommunication networks) drift may be the result of seasonality or periodicity effects (e.g., demands in water networks), software or hardware faults (e.g., sensor or actuator faults), and changes in user behaviour. Therefore, this requires new learning algorithms which are able to learn as data are observed one-by-one, and to continually adapt in order to maintain their effectiveness under drifting dynamics.

While data stream mining focuses on the arising challenges by dynamic data streams, the emergence of interconnected and complex systems requires a further evolution in our data analysis methods. For example, in the area of critical infrastructure systems \cite{kyriakides2014intelligent}, as the demand for their services is growing rapidly, these systems become larger, more complex and have evolved over time to become more heterogeneous and distributed. This has created interdependencies between them, with fault effects propagating to interdependent infrastructures. Other real-world examples include recommender systems in e-commerce, molecular graphs for drug discovery in chemistry, citation networks in academia, and social networks \cite{wu2020comprehensive}.

The intricate relationships and dependencies inherent in many real-world systems, are best modelled through graph structures, which encapsulate entities as nodes and relationships as edges \cite{wu2020comprehensive}. Graph stream mining extends the principles of data stream mining to dynamic graph structures, and aims at discovering insights and unraveling hidden patterns within evolving graph structures. This becomes a necessity for understanding the dynamics of graph structures and facilitating informed decision-making.

The contributions of this work are as follows. We consider the general setting where a data generating process produces a graph at each time of a variable number of nodes and edges, and propose a new method which uses incremental learning to continually update a model to perform graph classification. To achieve this, the method selects representative graphs per class, referred to as prototypes, which are then used to create graph embeddings; i.e., it embeds graphs into vectors. Furthermore, we complement incremental learning with a loss-based concept drift detection mechanism which re-calculates the graph prototypes when drift is detected.

The rest of the paper is structured as follows. Section~\ref{sec:formulation} provides the relevant definitions and formulates the problem of graph stream classification. Related work is discussed in Section~\ref{sec:related}. The proposed method is presented in Section~\ref{sec:method}. Section~\ref{sec:exp_setup} describes the experimental setup, while Section~\ref{sec:exp_results} discusses the experiments results. Section~\ref{sec:conclusion} concludes the paper and provides directions for future work.

\section{Definitions and Problem Formulation}\label{sec:formulation}
A \textbf{graph} $g$ is defined as $g = (V, E)$, where $V$ is the set of vertices or nodes, and $E$ is the set of edges \cite{wu2020comprehensive}. Let $v_i \in V$ be a node, and $e_{i, j} = (v_i, v_j) \in E, i \neq j$ be an edge pointing from $v_i$ and $v_j$. Let $A \in \mathbb{R}^{N \times N}$ denote the adjacency matrix with $A_{i, j} = 1$ if $e_{i, j} \in E$ and $A_{i, j} = 0$ if $e_{i, j} \notin E$. A graph can be directed where all edges are directed from one node to another, or undirected where any pair of connected nodes have inverse directions, i.e., $A = A^\intercal$.

An \textbf{attributed graph} \cite{wu2020comprehensive} refers to a graph which may have node and edge attributes or features. Let the number of nodes and edges in a graph be $|V| = N$ and $|E| = M$ respectively. Let $X_V \in \mathbb{R}^{N \times d}$ be the node feature matrix, with $x_v \in \mathbb{R}^d$ being a $d$-dimensional feature vector of node $v \in V$. Let $X_E \in \mathbb{R}^{M \times c}$ be the edge feature matrix, with $x_{e_{i, j}} \in \mathbb{R}^c$ being a $c$-dimensional feature vector of edge $(v_i, v_j) \in E$.  Additionally, a graph may also have global features referring to the overall graph. Attributed graphs enable the representation of richer information beyond structural relationships, thus facilitating the modeling of complex systems.

\textbf{Graph stream classification} addresses the challenges posed by continuous and dynamic sequences of attributed graphs. We consider a data generating process that provides a graph at each time step $t$ as follows: $D = \{(g^t, y^t)\}^\infty_{t=1}$, where the data are typically sampled from a potentially infinite sequence. The examples are drawn from an unknown probability distribution $p^t(g,y)$, where $g^t \in G$ is an attributed graph in the graph space $G$, and $y^t \in \{1, ..., K\}$ denotes the graph's class label (ground truth) in the target space $Y \subset \mathbb{Z}^+$, and $K \geq 2$ is the number of classes. Importantly, in this general framework, graphs can have a variable number of nodes and edges, and can be directed or undirected, thus covering a vast number of real-world applications.

A graph classifier observes a new graph $g^t$ at time $t$ and makes a prediction $\hat{y}^t$ based on a concept $h: G \rightarrow Y$. The classifier receives the true label $y^t$, its performance is evaluated using a loss function and is then trained based on the loss incurred. This process is repeated at each step. The gradual adaptation of the classifier without complete re-training $h^t = h^{t-1}.train(\cdot)$ is termed \textbf{incremental learning} \cite{losing2018incremental}.

Data \textbf{non-stationarity} \cite{ditzler2015learning} is a significant challenge in some streaming applications which is, typically, caused by \textbf{concept drift} \cite{gama2014survey, lu2018learning}, which represents a change in the joint probability distribution, defined as: $p^{t_i}(g, y) \neq p^{t_j}(g, y), i \neq j$.

\section{Related Work}\label{sec:related}

\subsection{Concept drift adaptation}
Typically, methods to address drift are grouped into active and passive \cite{ditzler2015learning}. Active methods employ explicit mechanisms to detect drift, such as statistical tests to detect changes in a data distribution (e.g., MannWhitney U test \cite{li2023autoencoder}), or threshold-based methods which monitor a performance metric (e.g., \cite{gama2004learning}). Typically, the model is completely retrained once drift is detected.

Alternatively, implicit methods exist which are referred to as passive methods, where they use incremental training to learn the drift \cite{elwell2011incremental} to continually update the classifier. Typically, they are memory-based, i.e., they have a memory (e.g., sliding windows) to store recent examples, which the model is incrementally trained on (e.g., \cite{malialis2020online}). Ensemble-based methods are also applied where a collection of classifiers exists, and members of the ensemble are added or discarded based on their performance (e.g., \cite{krawczyk2017ensemble}).

Other methods include hybrid methods which attempt to combine the strengths of both worlds (e.g., \cite{malialis2022hybrid}), while an alternative approach is called drift unlearning \cite{artelt2022unsupervised} which attempts to remove the effects / impact of drift on the data, and to revert the data distribution to the original (pre-drift).

\subsection{Graph stream classification}
Many methods have been proposed for traditional graph classification\footnote{Other graph tasks exist, such as node classification and link prediction, which are outside the scope of this work.} \cite{wu2020comprehensive, zhang2020deep, jin2023survey}, however, existing work on graph stream classification is limited. The survey in \cite{ranshous2015anomaly} provides a comprehensive overview of anomaly detection in dynamic networks. In \cite{barnett2016change}, change point detection has been examined in the context of correlation networks; a correlation network is a special type of network obtained by defining the edges based on some measure of correlation between each pair of nodes. The work in \cite{peel2015detecting} formalises the network change point detection problem in an online probabilistic learning framework, while the work in \cite{wilson2019modeling} focuses on network surveillance and proposes a block model for detecting a structural change.

\begin{figure*}[t!]
	\centering
	\includegraphics[scale=0.5]{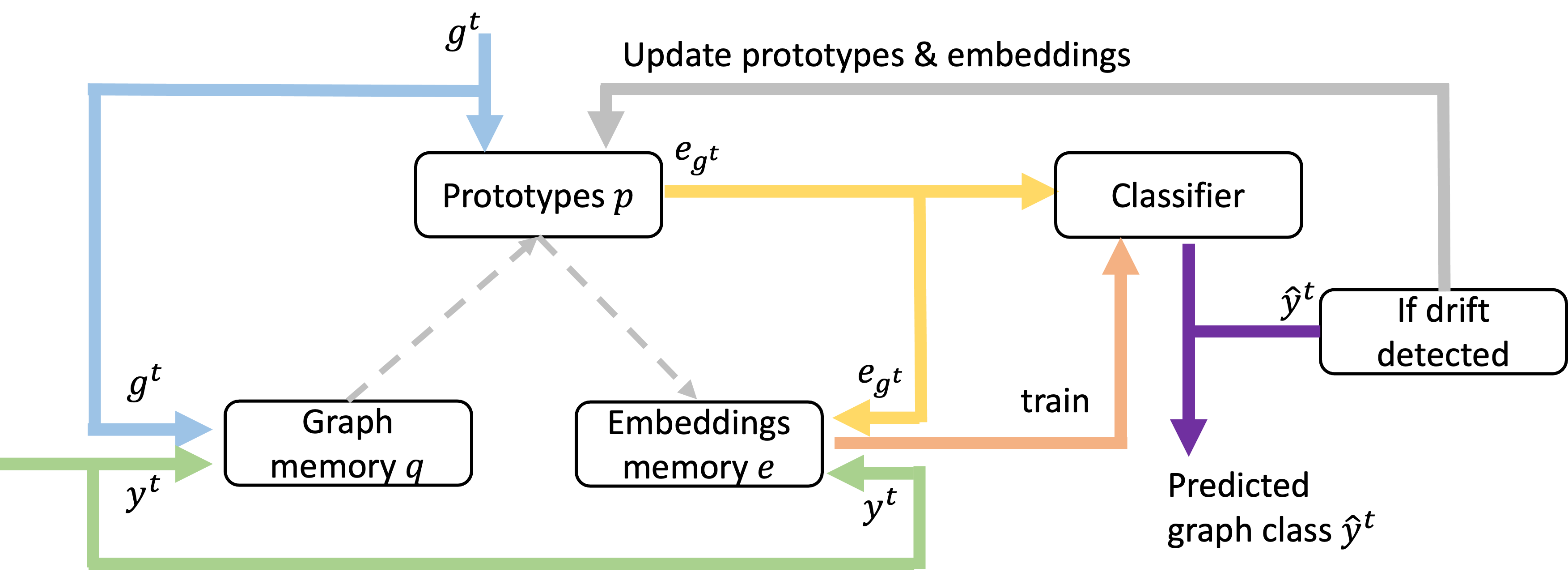}
	
	\caption{Overview of the proposed method.}
	
	\label{fig:overview}
\end{figure*}

The seminal paper \cite{zambon2018concept} constitutes the first work that uses graph embeddings for anomaly detection; it embeds each graph in the stream into a vector domain based on graph prototypes, thus enabling the use of a conventional change detection method to a more manageable setting. This paper \cite{grattarola2019change} adopts the idea of using graph embeddings, but instead, it learns the embeddings by training an autoencoder to represent graphs on constant-curvature Riemannian manifolds.

In our work, we adopt the use of prototype-based graph embeddings, but contrary to \cite{zambon2018concept, grattarola2019change}: (a) We depart from anomaly detection (normality vs abnormality) and propose a method for multi-class graph stream classification; (b) The proposed drift detection method is loss-based, in contrast to previous works where drift and anomaly are considered the same; (c) Our proposed method is hybrid as it proposes a synergy between incremental learning and drift detection.

\section{Method for Graph Stream Classification}\label{sec:method}
The overview of the proposed method is shown in Fig.~\ref{fig:overview}. A new graph $g^t$ is observed at time $t$, while a graph memory $q$ is maintained to store recent graphs (this is displayed in blue colour). The method also considers class prototypes $p$, i.e., representative graphs per class. Using the prototypes, the embedding $e_{g^t}$ of graph $g^t$ is calculated which is stored in a graph embeddings memory $e$ (yellow). Given the embedding, a classifier outputs a prediction $\hat{y}^t$ (purple). A concept drift detector is in place which updates the graph prototypes and embeddings if drift is detected (grey). The true label $y^t$ is provided (green, and the pairs $(g^t, y^t)$ and $(e_{g^t}, y^t)$ are appended to the graph memory and embeddings memory respectively. Lastly, incremental training is performed to update the classifier (orange). The pseudocode of the proposed method is provided in Alg.~\ref{alg:method}. A detailed description of each component in Fig.~\ref{fig:overview} and each line in Alg.~\ref{alg:method} is presented below.

\subsection{Incremental learning}

\begin{figure}[b!]
	\centering
	\includegraphics[scale=0.4]{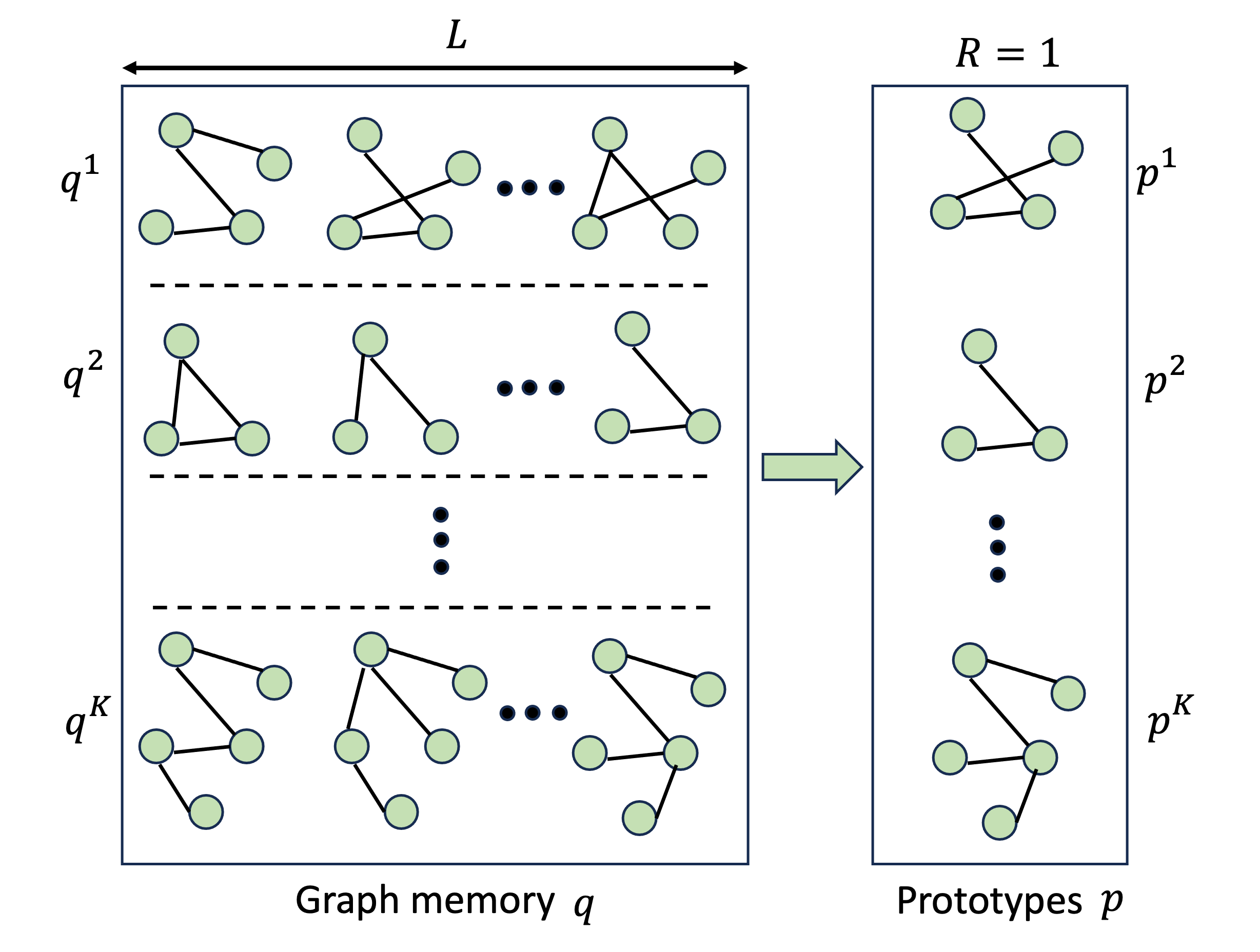}
	
	\caption{Class prototypes from graph memory.}
	
	\label{fig:prototypes}
\end{figure}

\textbf{Graph Memory}. 
The proposed method has multiple queues, one for each class, which is populated by arriving graphs as follows:
\begin{equation}\label{eq:memory}
	q = \{q^c\}^K_{c=1},
\end{equation}
\noindent where $K \geq 2$ is the number of classes, and $q^c$ is the queue corresponding to class $c$. Each queue is defined as follows:
\begin{equation}\label{eq:memory_class}
	q^c = \{g_i\}^L_{i=1},
\end{equation}
\noindent where $L$ is the size of each queue, and for any two $g_i, x_j \in q^c$ such that $j > i$, $g_j$ has been observed more recently in time. This is shown on the left-hand side of Fig.~\ref{fig:prototypes}.

\textbf{Graph prototypes}. For each class $c$ we calculate its prototype set $p^c$, that is, its representative graphs:
\begin{equation}
	p^c = \{p_i\}^R_{i=1},
\end{equation}
\noindent where $p_i \in q^c$ and $1 \leq R < L$ is the number of prototypes or representative graphs per class. The whole prototype set is then defined as follows:
\begin{equation}\label{eq:prototypes}	
	p = \{p^c\}^K_{c=1},
\end{equation}

This is shown on the right-hand side of Fig.~\ref{fig:prototypes}. To obtain the class prototypes $p^c$ we use the Centers algorithm \cite{riesen2007graph}, which assumes that the median graph in $q^c$ is the central graph. This is defined as the graph whose sum of distances to all other graphs is minimal:
\begin{equation}	
	p^c = med(q^c) = \argmin_{g_1 \in q^c} \sum_{g_2 \in q^c} \delta(g_1, g_2),
\end{equation}
\noindent where $\delta(\cdot, \cdot)$ is a distance metric between two graphs. To select $R$ prototypes per class we repeat the procedure but each time we exclude the already selected median graphs.

\textbf{Graph embeddings}. We adopt the embedding method from \cite{riesen2007graph} which was also adopted in \cite{zambon2018concept} for anomaly detection in graph streams. Given a graph $g \in G$ and all prototypes $p$, we calculate its embedding $e_g \in \mathbb{R}^{R \times K}$ as follows:
\begin{equation}\label{eq:embedding}
e_g = \{\delta(g, p_i)\}_{i=1}^{R \times K},
\end{equation}
\noindent where $p_i \in p$. Using this process, we convert all the graphs in the memory to their embedding form, analogous to Eqs. (\ref{eq:memory}) and (\ref{eq:memory_class}), as follows:
\begin{equation}\label{eq:embedding_memory}
\begin{aligned}
	e &= \{e^c\}^K_{c=1}\\
	e^c &= \{e_{g_i}\}^L_{i=1}
\end{aligned}
\end{equation}

\textbf{Incremental training}. A classifier $h$ observes a graph embedding $e_g \in \mathbb{R}^{R \times K}$ and makes a prediction $\hat{y} \in Y, Y \in \{1, 2, ..., K\}$, that is, $h: \mathbb{R}^{R \times K} \rightarrow Y$. The training set consists of the embeddings $e$ of the graphs in $q$. The cost function is defined as follows:
\begin{equation}\label{eq:cost}
	C = \frac{1}{L \times K} \sum_{i=1}^{L \times K} l(y_i, h(e_{g_i}))
\end{equation}
\noindent where $h(e_{g_i})$ is the predicted class of graph $g_i$ with an embedding $e_{g_i}$, and $l(\cdot, \cdot)$ is a loss function, such as the categorical cross-entropy. At each time step $t$, the classifier is incrementally updated based on the cost incurred, that is:
\begin{equation}
	h^t = h^{t-1}.train(\cdot)
\end{equation}

\subsection{Concept drift detection}

\textbf{Error modelling}. The prediction or 0/1 score at time $t$ is defined as $\mathbb{I}_{y^t == \hat{y}^t} \in \{0, 1\}$, where $\mathbb{I}$ is the identify function, $y^t$ is the ground truth and $\hat{y}^t$ is the prediction. We maintain two queues, a reference queue $q_{ref}$ and a moving queue $q_{mov}$, where $W = |q_{ref}| = |q_{mov}|$. The moving queue stores the most recent prediction scores, which will be compared to older scores stored in the reference queue. The prediction scores are modeled as Binomial distributions  \cite{gama2004learning}. For example, the mean and standard deviation for the reference queue are defined as:
\begin{equation}
	\begin{aligned}
		&	\mu_{ref} = p_{ref}\\
		&	\sigma_{ref} = \sqrt{\frac{p_{ref} (1 - p_{ref})}{W}},
	\end{aligned}
\end{equation}

\noindent where $p_{ref} = \frac{1}{|q_{ref}|} \sum_{s \in q_{ref}} s$ is the estimated parameter of the Binomial distribution, $s$ being each 0/1 prediction score. The mean $\mu_{mov}$ and standard deviation $\sigma_{mov}$ for $q_{mov}$ are calculated in a similar fashion.

\begin{algorithm}[t!]
	\caption{Proposed method for graph stream classification}
	\label{alg:method}
	\begin{algorithmic}[1]
		
		\Statex $L$ (Memory size per class); $R$ (Number of prototypes per class);  $W$ (Window size for drift detection); $\beta$ (Threshold for drift detection).
		
		\For{each time step $t$}
			\State observe graph $g^t$
			\State get its embedding $e_{g^t}$ \Comment Eq. (\ref{eq:embedding})
			\State predict graph class $\hat{y}^t = h^t(e_{g^t})$
			\State get ground truth $y^t$
			\State append 0/1 prediction score to $q_{mov}$
			
			\State check for drift between $q_{mov}$ and $q_{ref}$ \Comment Eq. (\ref{eq:drift_alarm})
			\If{drift detected}
				\State re-calculate graph prototypes \Comment Eq. (\ref{eq:prototypes})
				\State update embeddings in graph memory \Comment Eq. (\ref{eq:embedding_memory})
			\EndIf

			\State increment. train model $h^t = h^{t-1}.train(\cdot)$ \Comment Eq. (\ref{eq:cost})
		\EndFor
		
	\end{algorithmic}
\end{algorithm}

\textbf{Distribution change}. To detect a change in the distribution of prediction scores and, specifically, a decline in the model's performance, we calculate the following threshold:
\begin{equation}\label{eq:thresholds}
	\begin{aligned}
		\theta_{ref} &= \mu_{ref} - \beta \sigma_{ref},
	\end{aligned}
\end{equation}
\noindent where $\beta > 0$ is a sensitivity parameter which is task-dependant. A concept drift alarm will be triggered when the average of recent prediciton scores fall below the above threshold, i.e., when the following condition is satisfied:
\begin{equation}\label{eq:drift_alarm}
    \mu_{mov} \leq \theta_{ref}
\end{equation}

\textbf{Re-calculation of graph prototypes}. When concept drift is detected the graph prototypes $p$ are re-calculated, and all the graph embeddings $e$ in the memory are replaced.

\section{Experimental Setup}\label{sec:exp_setup}
This section describes the datasets used in our experimental study, the compared methods, and the evaluation methodology.

\subsection{Datasets}

\begin{figure}[t!]
	\centering
	\includegraphics[scale=0.5]{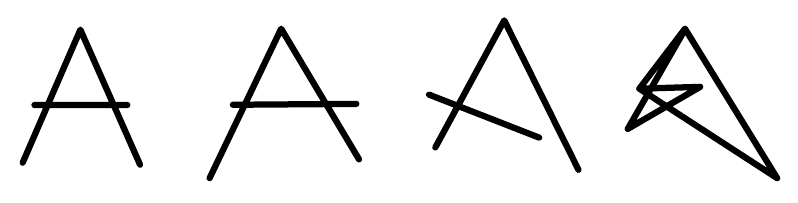}
	
	\caption{Distortions levels of character ``A'' in the Letter dataset.}
	
	\label{fig:letter}
\end{figure}

\textbf{Letter} \cite{riesen2008iam}. The dataset considers the three capital letters ``A'', ``I'' and ``Z'', which are converted into graphs by representing lines by undirected edges and ending points of lines by nodes. The graphs are uniformly distributed over the three classes. Each node is labelled with a two-dimensional attribute giving its position relative to a reference coordinate system. To test the compared methods under conditions of concept drift, distortions of different levels are applied on the graphs as shown in Fig.~\ref{fig:letter}. We consider two variants of the dataset, in which we vary the frequency and severity of concept drift:
\begin{itemize}
	\item \textbf{Letter\_high}. There are 300 arriving graphs without any distortions, followed by 450 with high distortion level.
	\item \textbf{Letter\_med\_high}. There are 300 arriving graphs without any distortions, followed by 450 with medium distortion level, and 450 with high distortion level.
\end{itemize}

\begin{figure}[t!]
	\centering
	\includegraphics[scale=0.8]{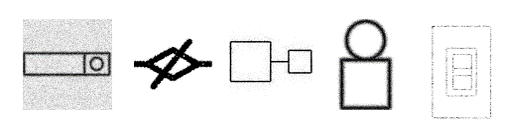}
	
	\caption{Distortion levels of GREC images.}
	
	\label{fig:GREC}
\end{figure}

\begin{figure}[t!]
	\centering
	
	\subfloat[left]{\includegraphics[scale=0.4]{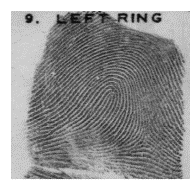}}%
		\label{fig:fingerprint_left}
	\subfloat[arch]{\includegraphics[scale=0.4]{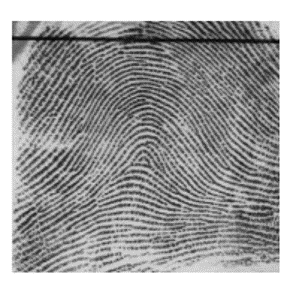}}%
		\label{fig:fingerprint_arch}

	\caption{Examples of the two classes in the Fingerprint dataset.}
 \label{fig:fingerprint}
\end{figure}

\textbf{GREC} \cite{riesen2008iam}. The GREC dataset comprises graphs representing symbols found in architectural and electronic drawings, with images at five distortion levels. In Fig.~\ref{fig:GREC}, each distortion level showcases a drawing example. Distortion levels determine the application of morphological operations, such as erosion or dilation. The resulting images are thinned to one-pixel-wide lines, from which graphs are extracted by tracing lines and detecting intersections and corners. Three classes are considered in this study. The graphs are uniformly distributed over the three classes. Nodes represent endpoints, corners, intersections, and circles, labeled with two-dimensional attributes indicating their positions. Undirected edges connect nodes, labeled as either lines or arcs.

\textbf{Fingerprint} \cite{riesen2008iam}. This dataset comprises two distinct fingerprint types, namely 'arch' and 'left', as shown in Fig.~\ref{fig:fingerprint}, which are transformed into graphs by nodes and edges. Each node is assigned a two-dimensional attribute representing its position, while the edges are characterised by an angle indicating their orientation relative to the horizontal direction. Derived from the NIST-4 reference database of fingerprints, this dataset features graphs that are evenly distributed across the specified classes.

For our implementation we have used NetworkX \cite{hagberg2008exploring}, a Python package for the creation, manipulation, and study of graphs. For the graph similarity / distance metric, we have used the Graph Edit Distance \cite{abu2015exact}, which is defined as the minimum cost of a series of edit operations (e.g., insert a link) required to transform one graph into another. 

\subsection{Compared methods}

\begin{table}[t!]
\caption{Hyper-parameter values for each dataset}\label{tab:params_nn}
\centering
\begin{tabular}{|c|ccc|}
\hline
                   & \multicolumn{1}{c|}{Letter}        & \multicolumn{1}{c|}{GREC}          & Fingerprint \\ \hline
Learning rate      & \multicolumn{1}{c|}{0.001}         & \multicolumn{1}{c|}{0.001}         & 0.01        \\ \hline
Hidden layers      & \multicolumn{1}{c|}{{[}128, 64{]}} & \multicolumn{1}{c|}{{[}128, 64{]}} & [32]          \\ \hline
Mini-batch size    & \multicolumn{1}{c|}{128}           & \multicolumn{1}{c|}{128}           & 128         \\ \hline
L2 regularisation      & \multicolumn{1}{c|}{0.0001}        & \multicolumn{1}{c|}{0.0001}        & 0.0           \\ \hline
Weight initialiser & \multicolumn{3}{c|}{He Normal \cite{he2015delving}}                                                        \\ \hline
Optimiser          & \multicolumn{3}{c|}{Adam \cite{kingma2014adam}}                                                             \\ \hline
Hidden activation  & \multicolumn{3}{c|}{Leaky ReLU \cite{maas2013rectifier}}                                                       \\ \hline
Num. epochs        & \multicolumn{1}{c|}{1}             & \multicolumn{1}{c|}{10}            & 1           \\ \hline
Output activation  & \multicolumn{3}{c|}{Softmax}                                                          \\ \hline
Loss function      & \multicolumn{3}{c|}{Categorical cross-entropy}                                             \\ \hline
\end{tabular}

\end{table}

\textbf{Proposed method}. This has been described in Section~\ref{sec:method} and its pseudocode is shown in Alg.~\ref{alg:method}.

\textbf{Feature-based method}. We implement the method from \cite{zambon2018concept}, which uses two features to represent graphs. The first feature of graph $g$ is the density of edges defined as: $\phi_1(g) = \frac{M}{N (N - 1)}$, where $N=|V|$ and $M=|E|$ are the number of nodes and edges respectively. The second feature is the spectral gap of graph $g$ defined as: $\phi_2(g) = |\lambda_1| - |\lambda_2|$, where $\lambda_1$ and $\lambda_2$ are the largest and second-largest eigenvalues of the graph's Laplacian matrix. The reasoning behind this particular choice is that both features are suitable for describing graphs with a variable number of nodes and edges.

For fairness, the graph memory from the proposed method is also used in the feature-based method to store the features (instead of the embeddings).
Additionally, for both methods the memory is assumed to be initially full by limited historical data (e.g., $L=10$ graphs per class). No pre-training takes place as learning starts at time $t=1$.

Both methods use the same classifier, a standard fully-connected neural network whose hyper-parameters for each dataset are shown in Table~\ref{tab:params_nn}. This is implemented in Keras \cite{chollet2015keras} with the PyTorch backend \cite{paszke2019pytorch}.

\subsection{Evaluation methodology}
We use the popular performance metric geometric mean \cite{sun2006boosting}, which is suitable for imbalanced datasets, and is defined as follows:
\begin{equation}\label{eq:gmean}
	G\text{-}mean = \displaystyle\sqrt[K]{\prod_{c=1}^K r_c},
\end{equation}
\noindent where $r_c$ is the recall of class $c$ and $K$ is the number of classes. It has the desirable property that it is high when all recalls are high and when their difference is small \cite{he2008learning}.

Additionally, we use the widely adopted prequential evaluation with fading factors method \cite{gama2013evaluating}, which has two major advantages. First, it has been proven to converge to the Bayes error when learning in stationary data. Second, it does not require a holdout set and the predictive algorithm is always tested on unseen data. The fading factor is set to 0.99. In all simulations, we plot the prequential G-mean in every step averaged over 10 repetitions, including the error bars displaying the standard error around the mean.

\section{Experimental Results}\label{sec:exp_results}

\begin{figure}[t!]
	\centering
	\includegraphics[scale=0.2]{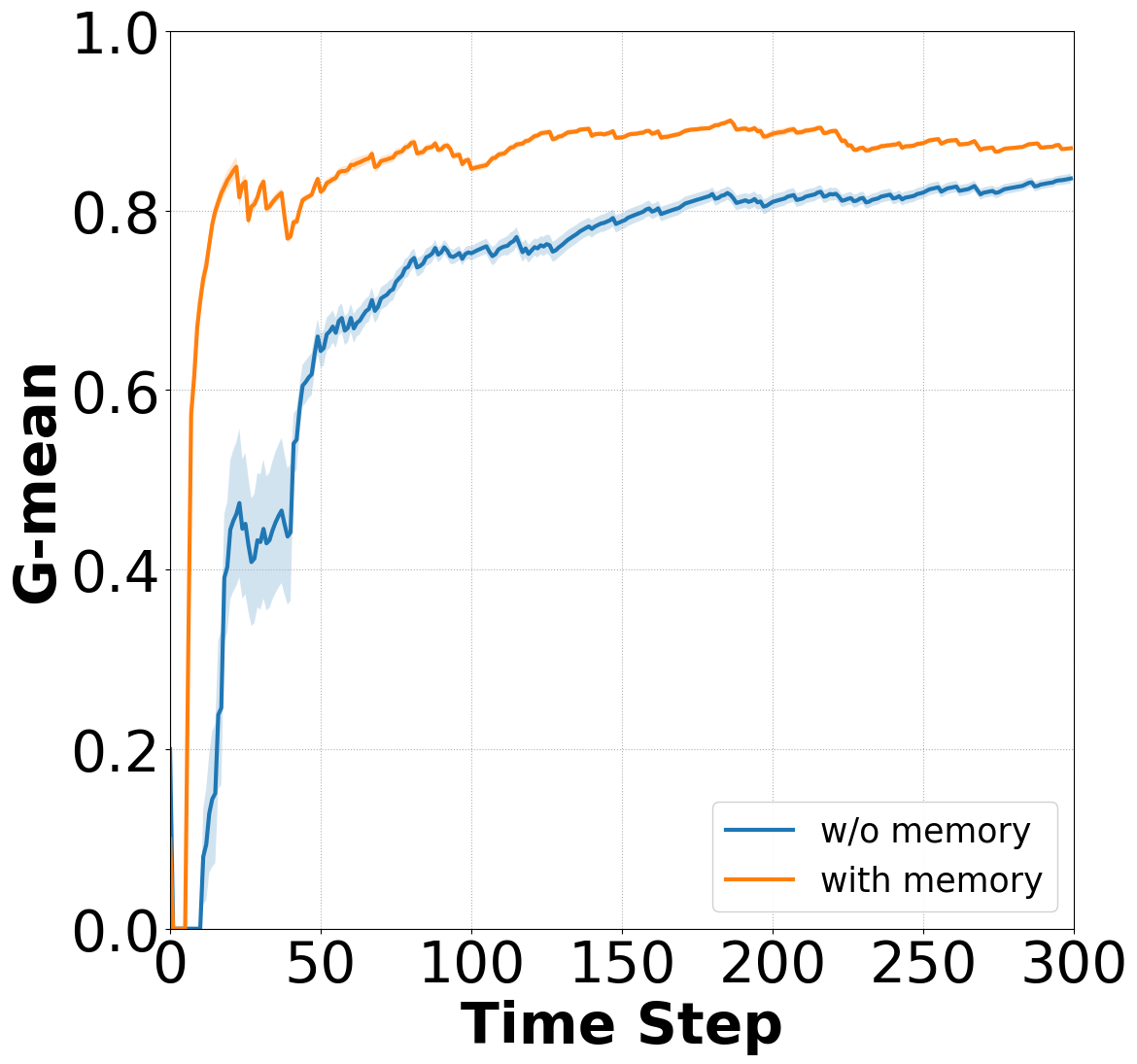}
	
	\caption{Performance of the proposed method with and without graph memory.}
	
	\label{fig:mem_letter3}
\end{figure}

\textbf{Role of the graph memory}. In this study, we will inspect the role of the graph memory in the proposed method. The result for Letter (without drift) is shown in Fig.~\ref{fig:mem_letter3}, where using the graph memory significantly outperforms the proposed method without memory. The main benefit of the graph memory concerns the learning speed as towards the end, they end up with a small performance different. Another major advantage of using the memory is that it would allow us to re-calculate the class prototypes when concept drift is detected. For a fair assessment, we have chosen the Letter variant without drift in this experiment. From now on, the graph memory will be used.

\begin{figure}[t!]
	\centering
	
	\subfloat[Letter\_high]{\includegraphics[scale=0.15]{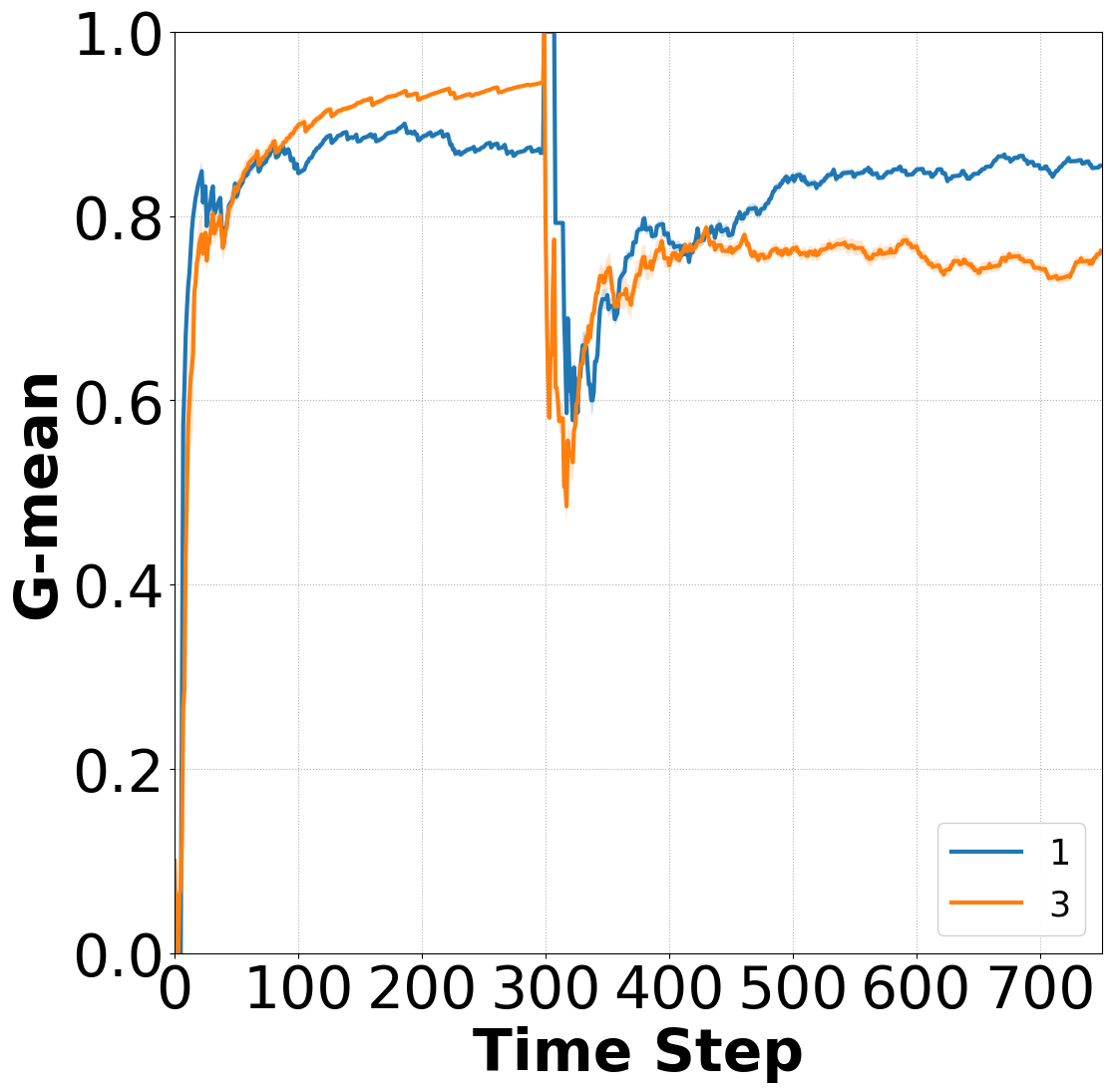}%
		\label{fig:prot_letter3_drift_high}}
	\subfloat[Letter\_med\_high]{\includegraphics[scale=0.15]{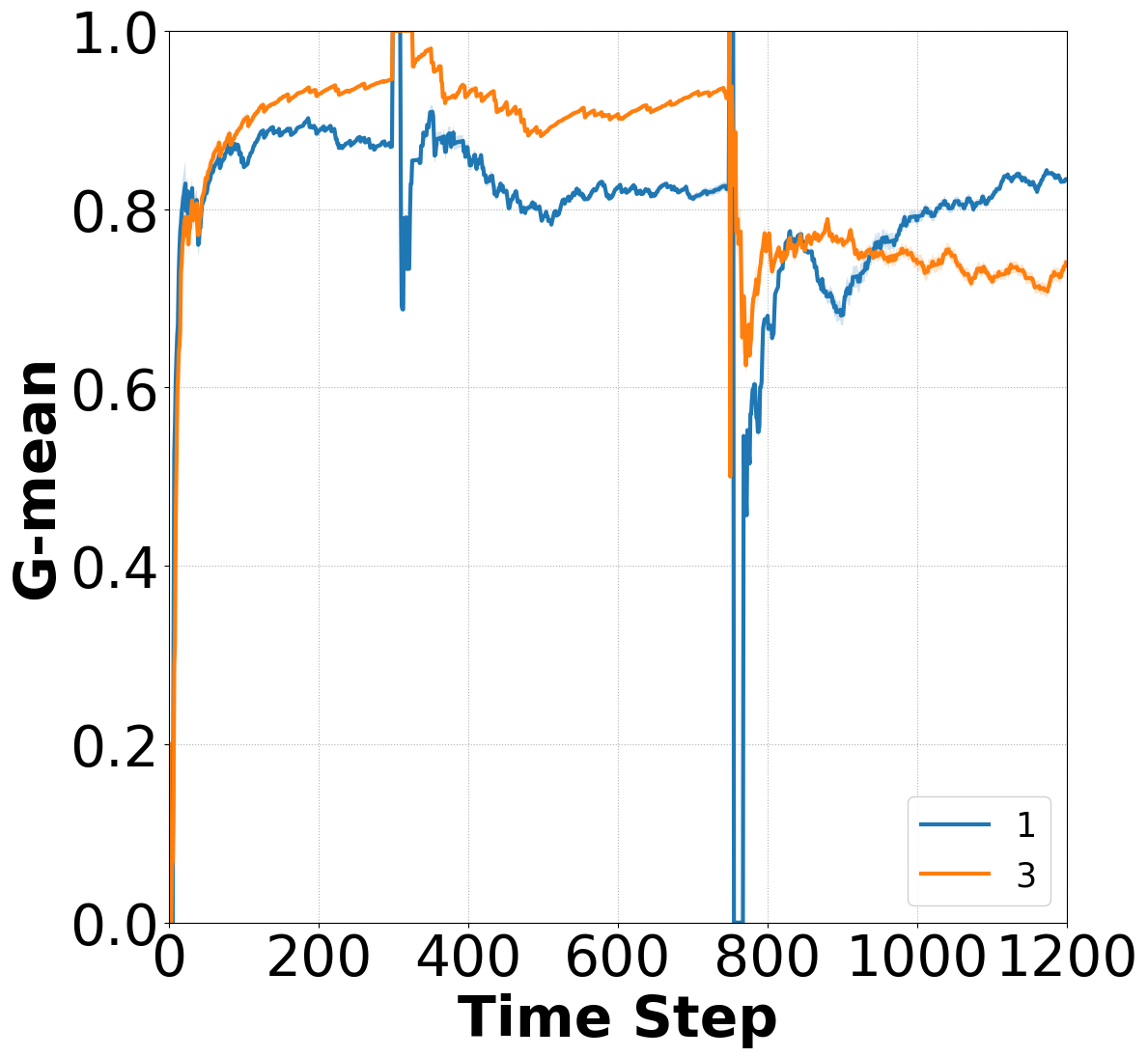}%
		\label{fig:prot_letter3_drift_med_high}}

    \subfloat[GREC]{\includegraphics[scale=0.15]{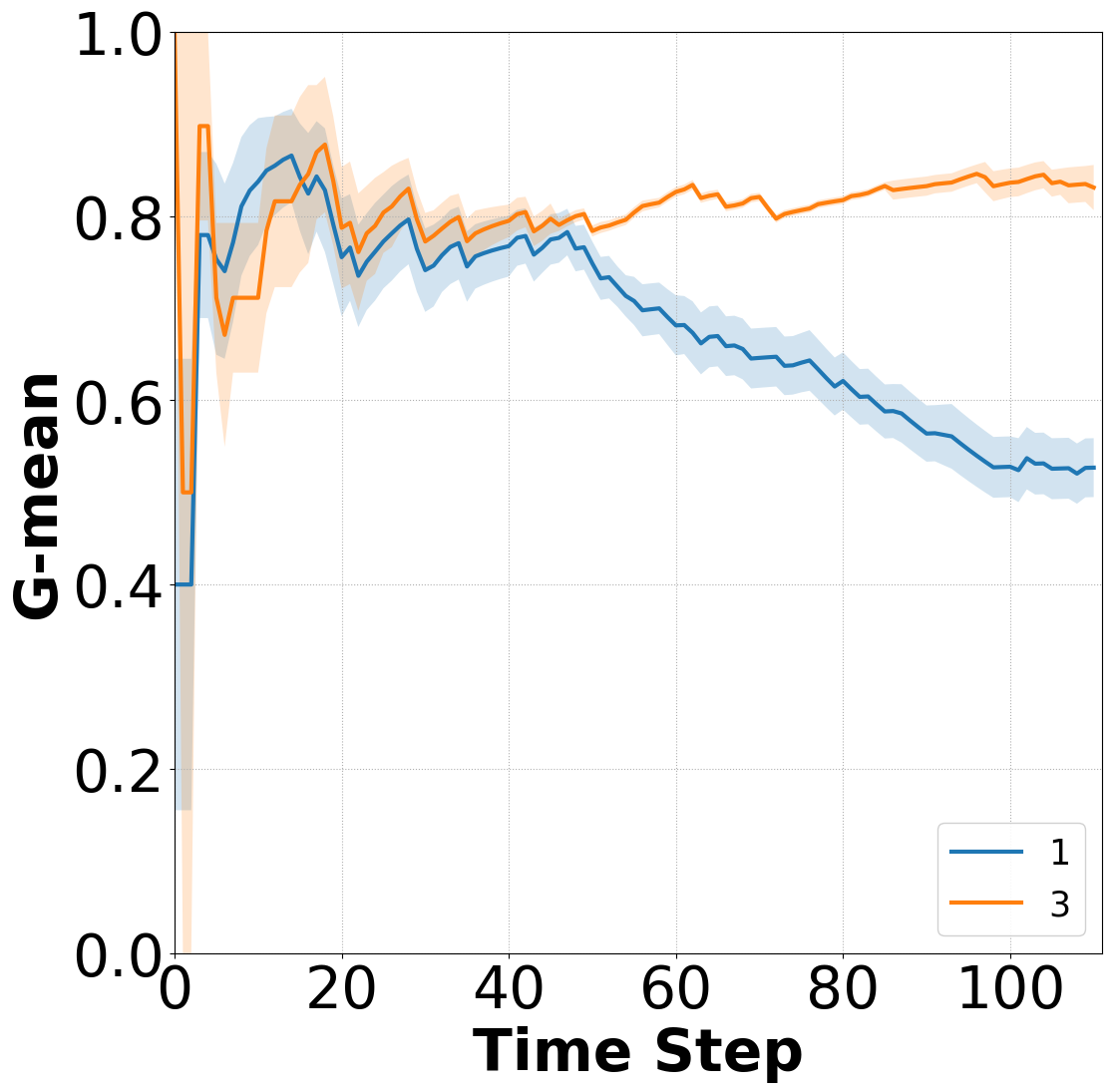}%
		\label{fig:prot_grec_1_3_compare_10}}
    \subfloat[Fingerprint]{\includegraphics[scale=0.15]{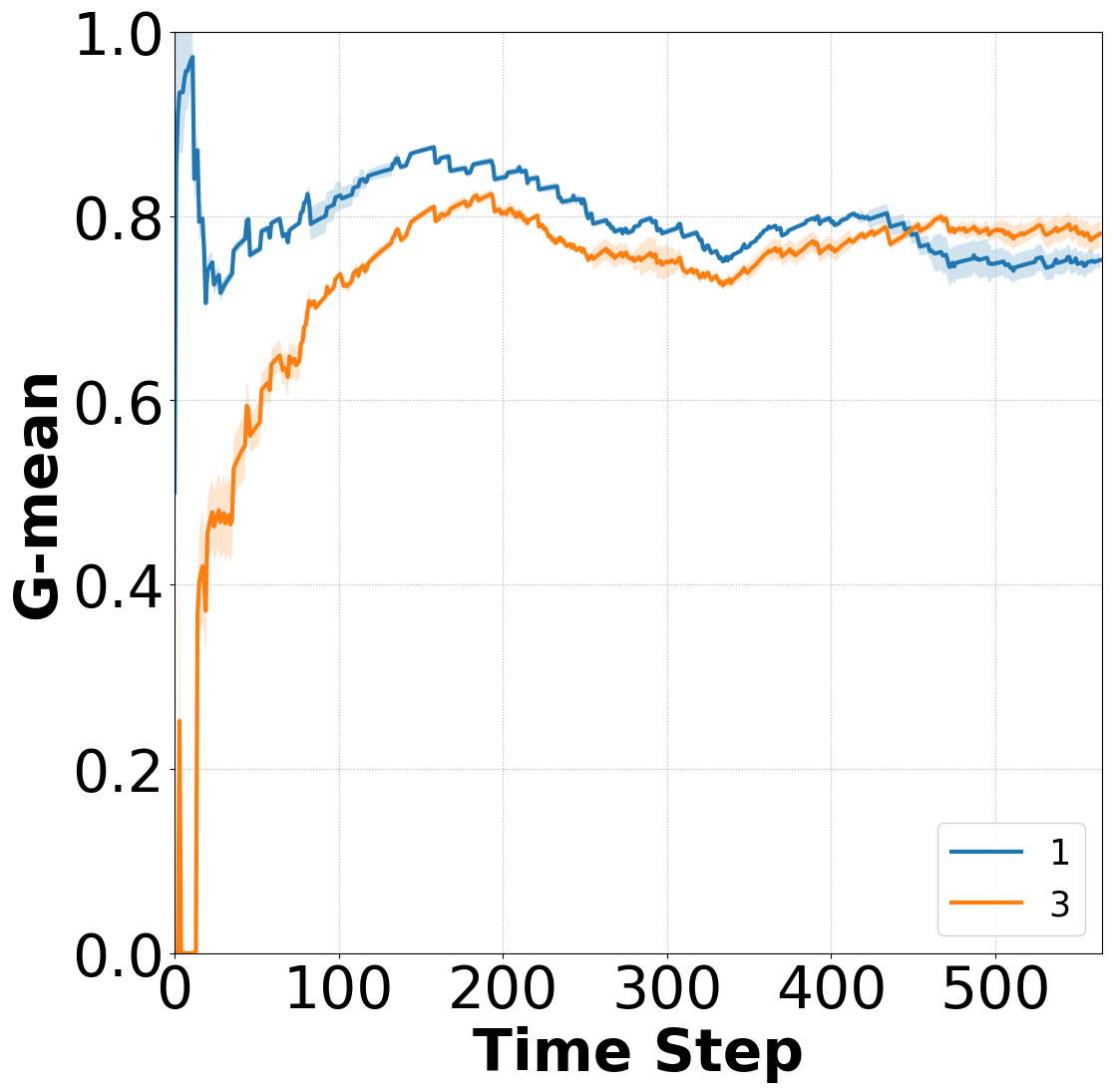}%
		\label{fig:prot_fingerprint_1_3_compare_10}}
	
	\caption{The performance of the proposed method with different number of prototypes ($R = 1, 3$) per class.}
 
	\label{fig:prototypes_result}
\end{figure}

\textbf{Number of class prototypes}. In this study, we examine the impact of the number of class prototypes on the model's performance. The memory size per class for all datasets is $L = 10$. As depicted in Fig.~\ref{fig:prot_letter3_drift_high} and Fig.~\ref{fig:prot_letter3_drift_med_high}, it is evident that the model's performance is superior with three prototypes compared to using only one when there is no concept drift or when drift is mild. After severe concept drift, the performance deteriorates indicating that old prototypes may no longer be accurate class representatives, therefore, a smaller number of them is preferred. In Fig.~\ref{fig:prot_grec_1_3_compare_10}, it is evident that the model's performance significantly improves when employing three prototypes as opposed to just one. In Fig.~\ref{fig:prot_fingerprint_1_3_compare_10}, although the model's performance with three prototypes initially lags behind that of one prototype, it steadily improves over time and ultimately slightly surpasses the latter. These findings collectively affirm that in scenarios with limited or without drift, the model consistently exhibits superior performance when using three prototypes as opposed to relying on only one.

\begin{figure}[t!]
	\centering
	
	\subfloat[Letter\_high]{\includegraphics[scale=0.145]{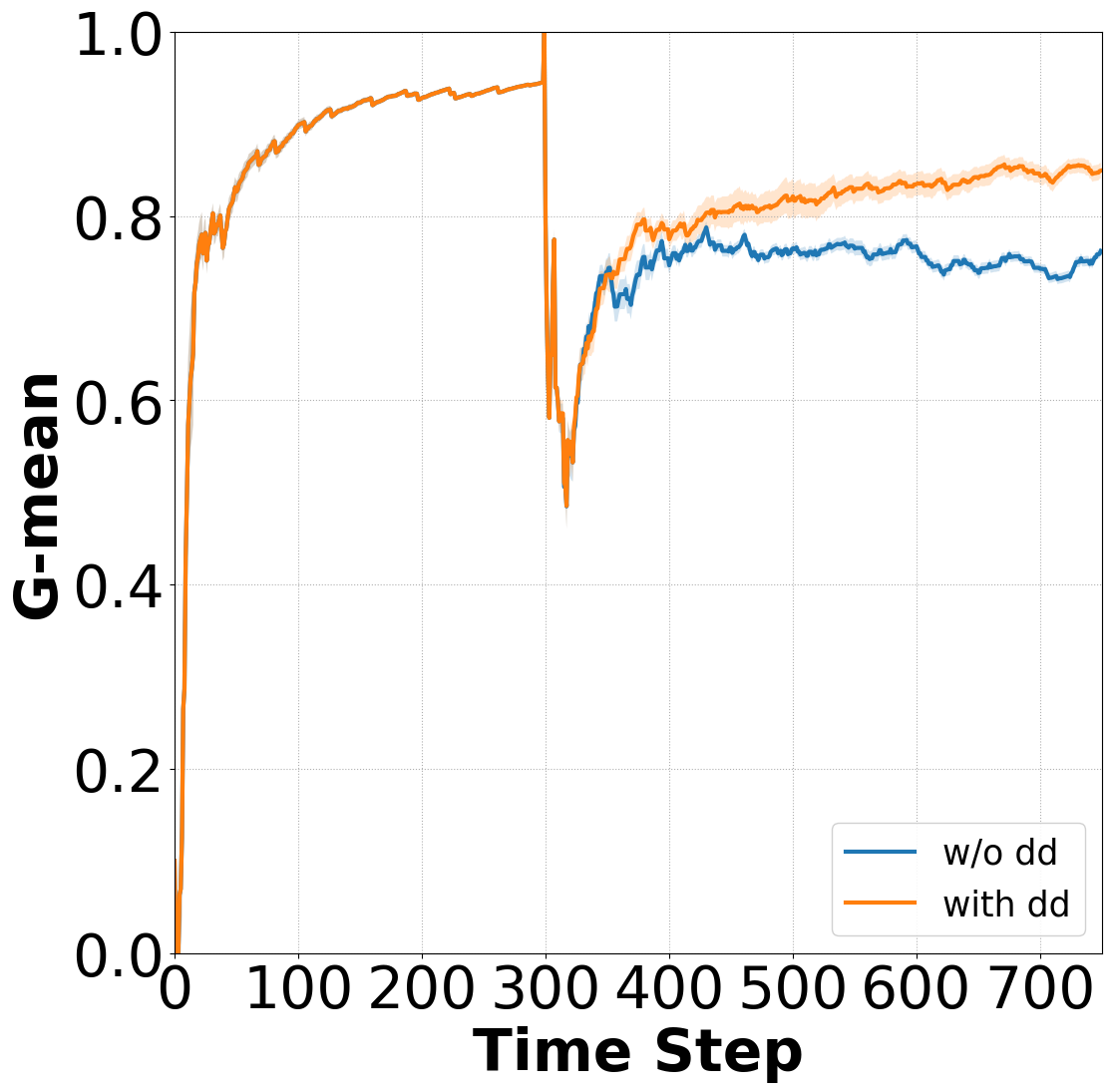}}%
		\label{fig:letter_high_drift}
	\subfloat[Letter\_med\_high]{\includegraphics[scale=0.145]{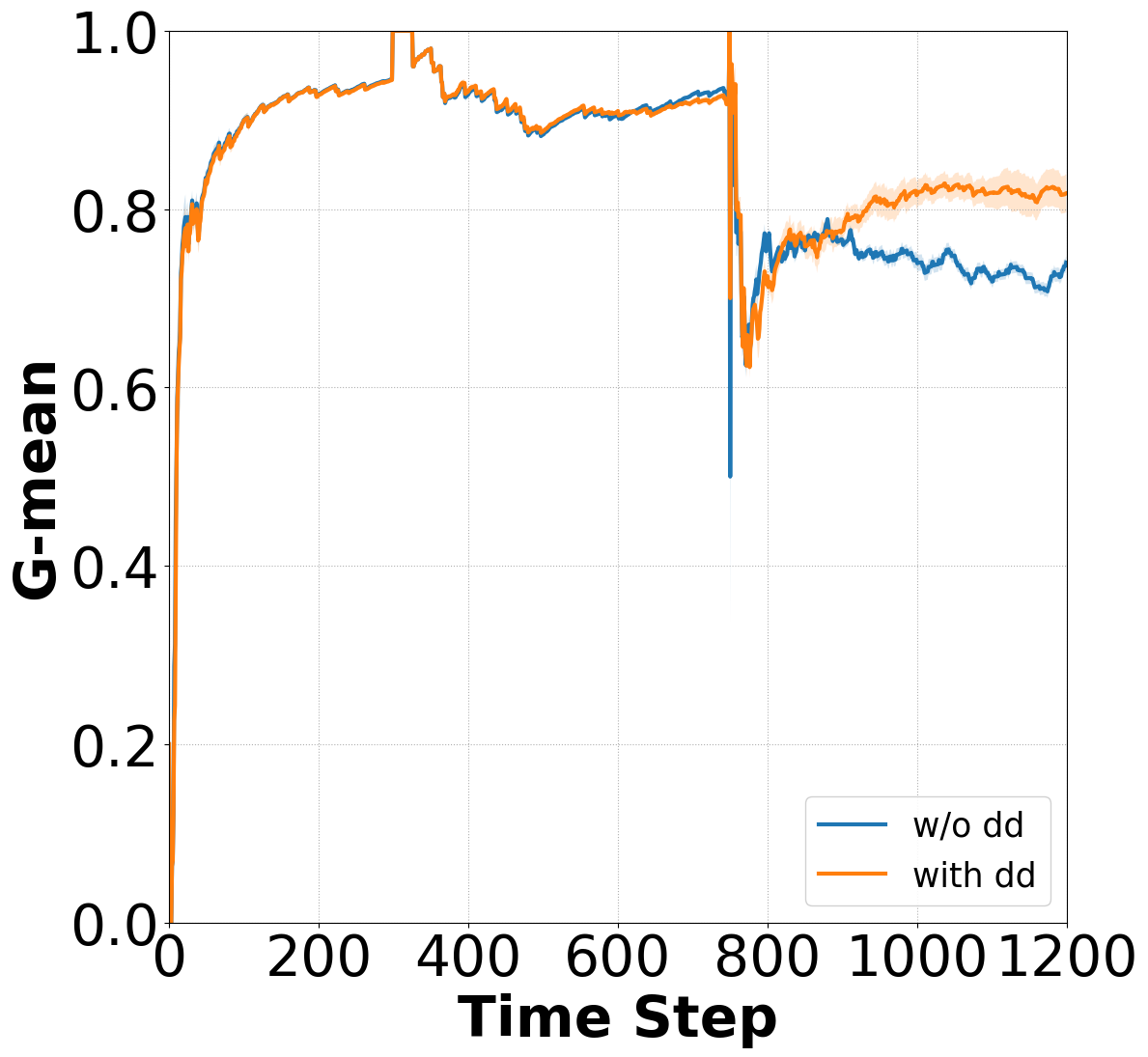}}%
		\label{fig:letter_med_high_drift}

	\caption{The performance of the proposed method with and without concept drift detection.}
 \label{fig:drift_result}
\end{figure}

\textbf{Concept drift detection}. Here, we examine the impact of the drift detector on the model's performance. The experiment is carried out with a memory size of $L=10$, $R=3$ prototypes per class, and $\beta = 4.5$ for the drift detection sensitivity parameter. As depicted in Fig.~\ref{fig:drift_result}, the model's performance is similar in both cases regardless of the presence of the drift detector. However, following a drift occurrence, there is a noticeable improvement in performance for the model equipped with the drift detector. This provides empirical evidence supporting the assertion that a drift detection mechanism can effectively enhance overall model performance. This is attributed to the fact that the graph prototypes are re-calculated, thus replacing the old ``obsolete'' ones.

\begin{figure}[t!]
	\centering
	
	\subfloat[Letter\_high]{\includegraphics[scale=0.145]{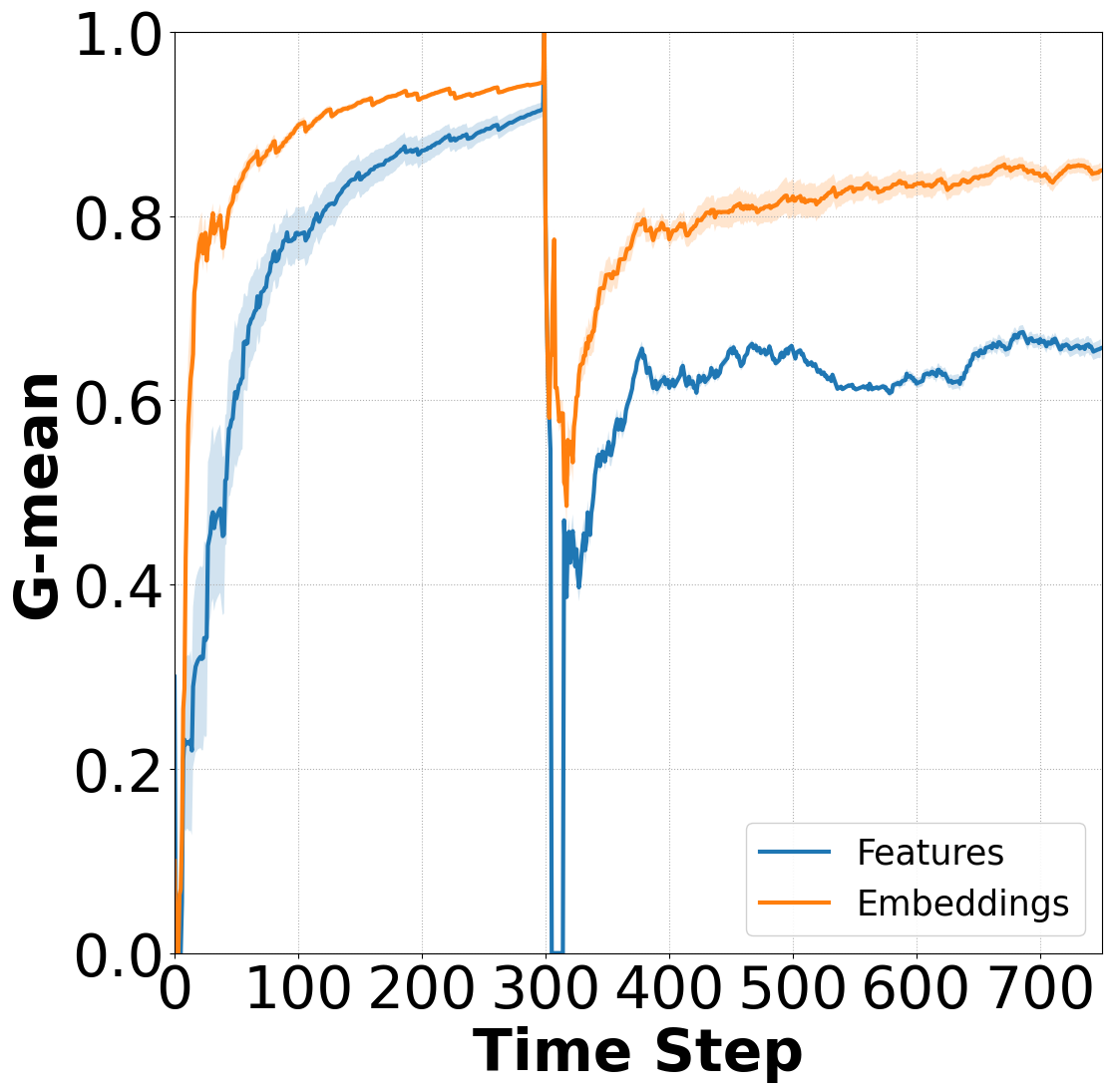}}%
		\label{fig:comparison_letter3_drift_high}
	\subfloat[Letter\_med\_high]{\includegraphics[scale=0.145]{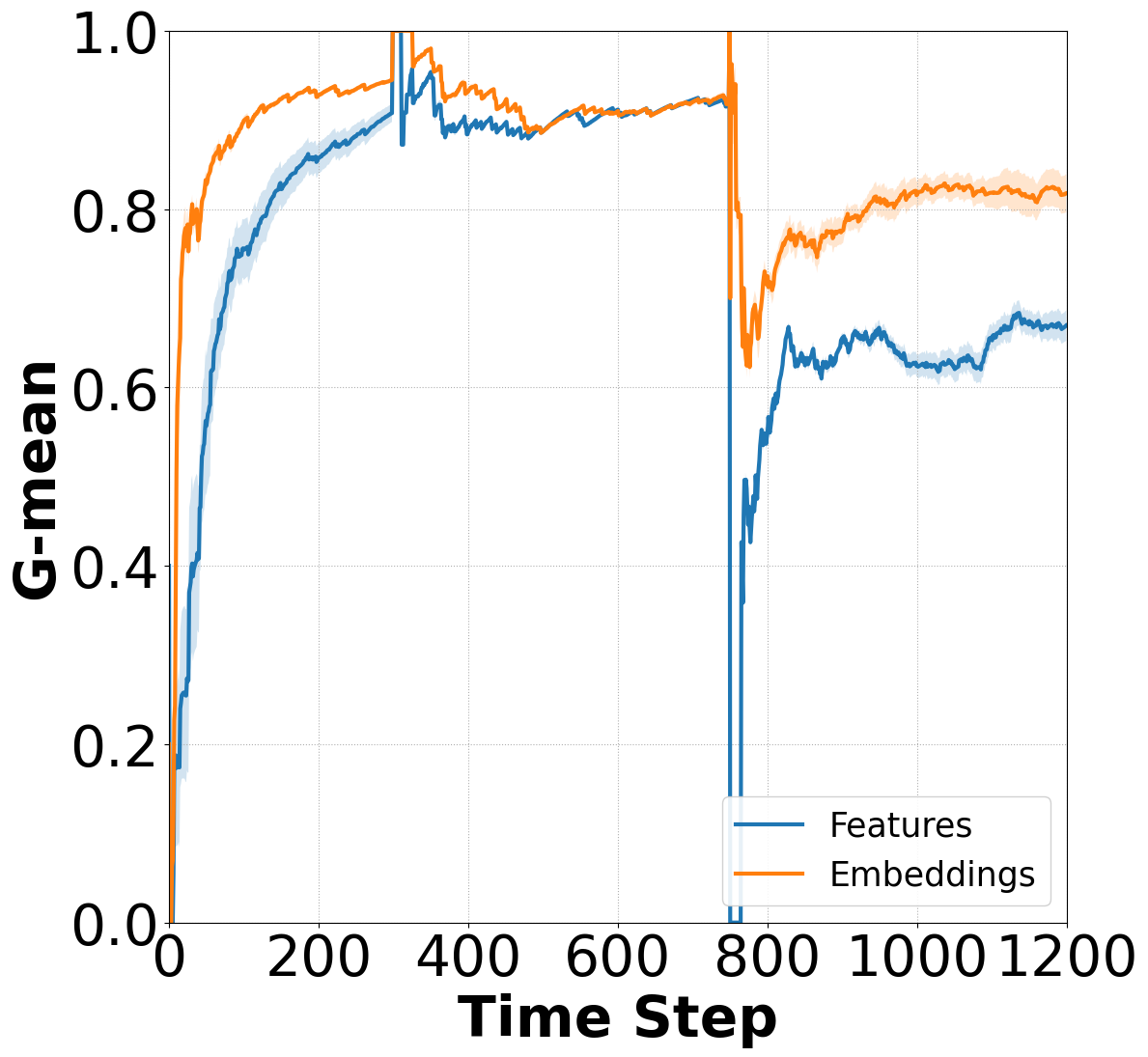}}%
		\label{fig:comparison_letter_drift_med_high}
  \subfloat[GREC]{\includegraphics[scale=0.145]{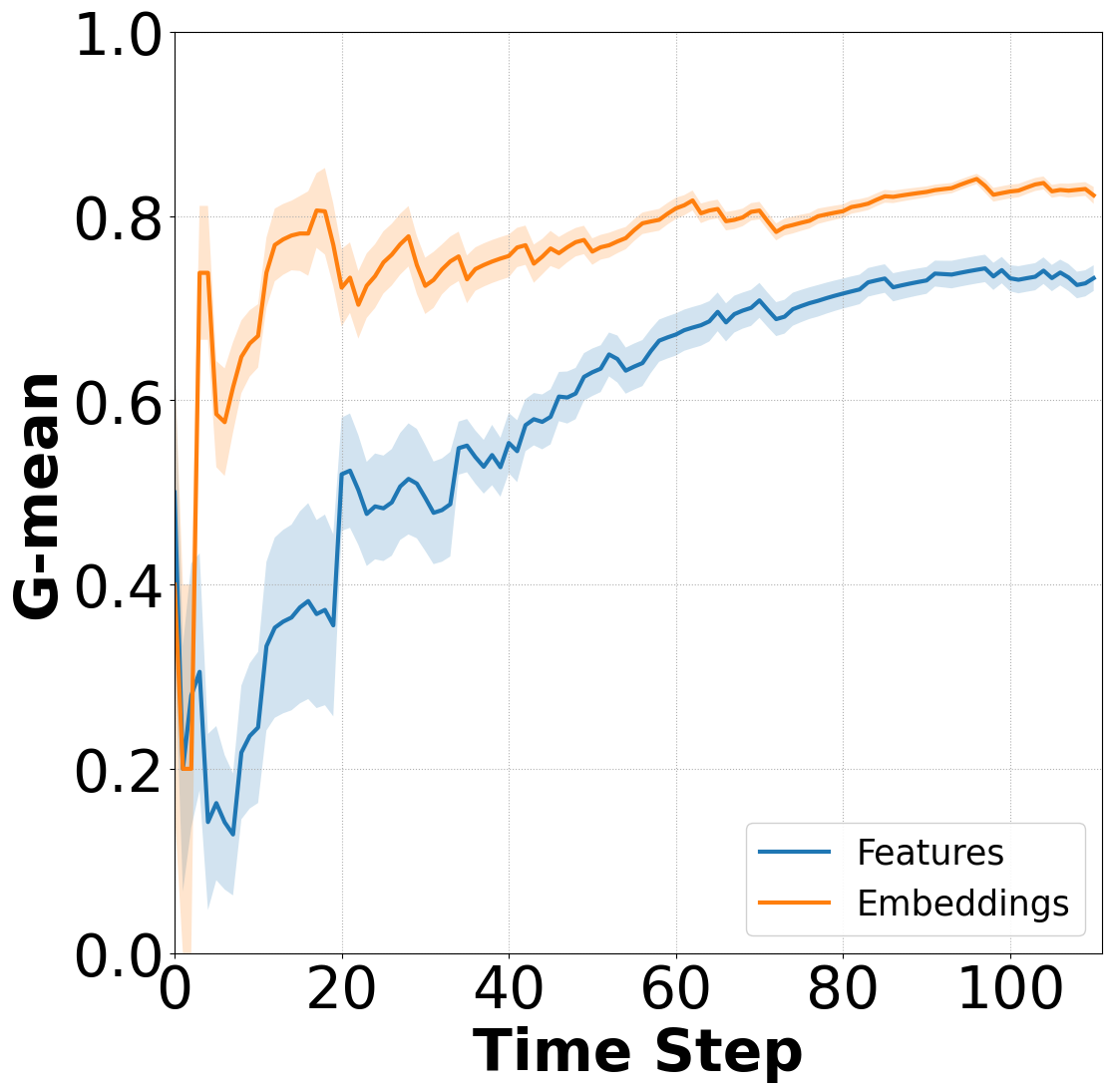}}%
		\label{fig:comparison_grec}
   \subfloat[Fingerprint]{\includegraphics[scale=0.145]{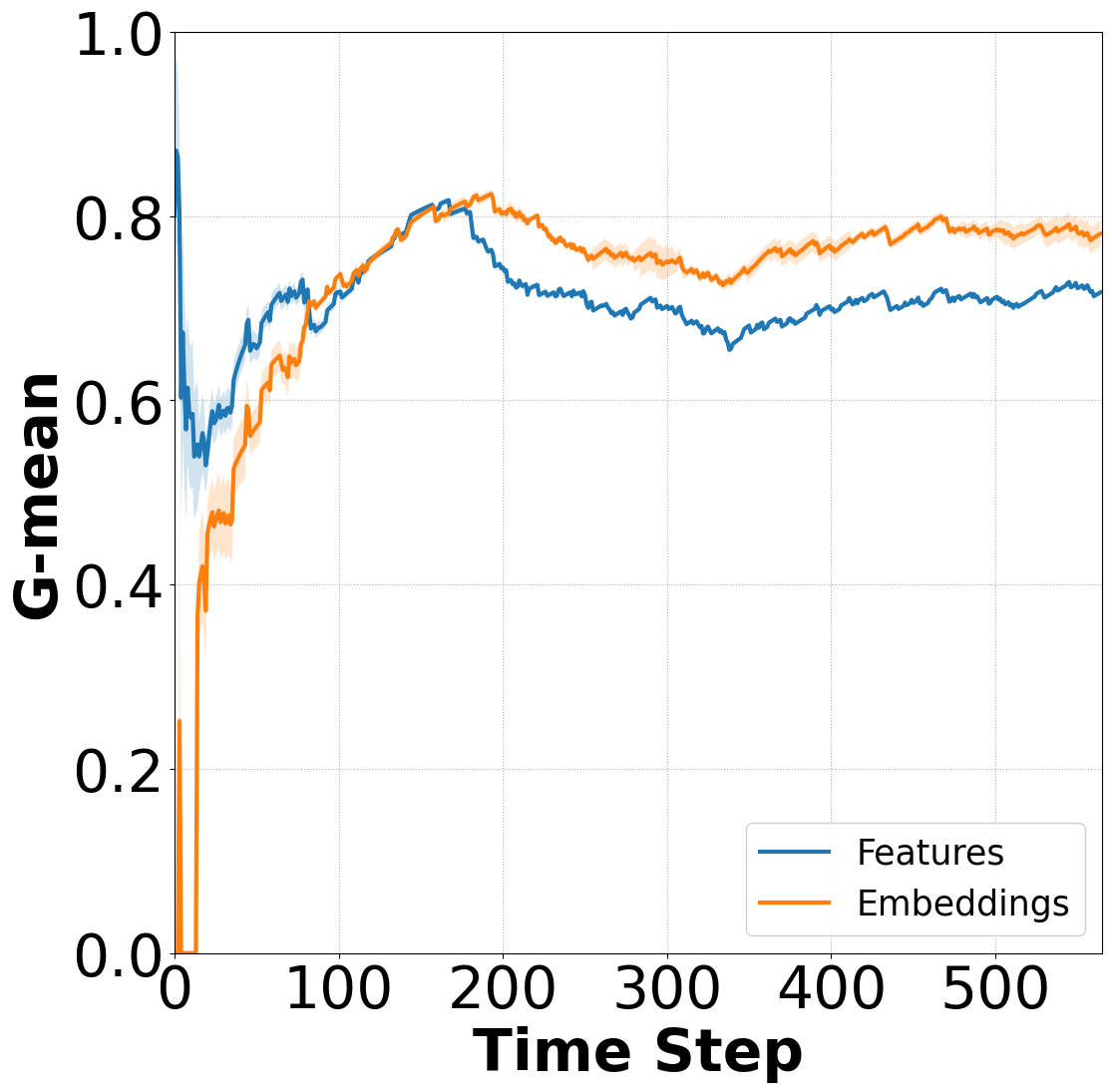}}%
		\label{fig:comparison_fingerprint}

	\caption{Comparison between the proposed method (using embeddings) and the feature-based method.}
 \label{fig:comparison_result}
\end{figure}

\textbf{Comparative study}. In this section, we present a comparative analysis of our proposed method (which uses graph embeddings) against the feature-based method. The outcomes are shown in Fig.~\ref{fig:comparison_result}. A significant performance difference between the two methods is evident across all datasets. Our proposed method consistently surpasses the feature-based, both prior to and following the occurrence of drift in the Letter dataset. Also, our method performs better in the GREC and Fingerprint datasets.

\section{Conclusion and Future Work}\label{sec:conclusion}
This work introduces a novel method for graph stream classification which operates under the general setting where graphs have a variable number of nodes and edges. It proposes the use of incremental learning for continual model adaptation using class prototype-based graph embeddings, in conjunction with a loss-based concept drift detection mechanism. Directions for future work are provided below.

\textbf{Learning graph embeddings}. In this work, the graph embeddings were calculated based on graph prototypes. Future work will examine methods to learn the embeddings (e.g., \cite{grattarola2019change}), which we will allow us to apply the proposed method to large-scale graph stream problems.

\textbf{Limited ground truth availability}. In this work, we have assumed the availability of the graph class label after each prediction step. Future work will consider other learning paradigms, such as unsupervised \cite{li2023autoencoder}, semi-supervised \cite{pan2013graph} and active \cite{zliobaite2013active} learning, and methods, such as few-shot learning \cite{malialis2022nonstationary} and data augmentation \cite{malialis2022data}; all of which have been shown to be effective in traditional data stream mining.

\bibliographystyle{IEEEtran}
\bibliography{paper}

\end{document}